\documentclass{ifacconf}

\usepackage{graphicx}      
\usepackage{natbib}        
\usepackage[acronym]{glossaries} 
\usepackage{fancyhdr}

\newacronym{llm}{LLM}{Large Language Model}
\newacronym{mcp}{MCP}{Model Context Protocol}

\usepackage{url}

\begin{document}

\begin{frontmatter}

\title{Benchmark for Planning and Control with Large Language Model Agents: Blocksworld with Model Context Protocol \thanksref{footnoteinfo}} 

\thanks[footnoteinfo]{This work has been submitted to IFAC for possible publication}

\author[First]{Niklas Jobs} \author[First]{Luis Miguel Vieira da Silva} \author[First]{Jayanth Somashekaraiah} \author[First]{Maximilian Weigand} \author[Second]{David Kube} \author[First]{Felix Gehlhoff}

\address[First]{Institute of Automation Technology, Helmut Schmidt University / University of the Federal Armed Forces Hamburg, Germany \\
(e-mail: \{firstname.lastname\}@hsu.hamburg)}
\address[Second]{Siemens AG, Nuremberg, Germany (e-mail: david.kube@siemens.com)}

\begin{abstract}                
Industrial automation increasingly requires flexible control strategies that can adapt to changing tasks and environments. Agents based on Large Language Models (LLMs) offer potential for such adaptive planning and execution but lack standardized benchmarks for systematic comparison. We introduce a benchmark with an executable simulation environment representing the Blocksworld problem providing five complexity categories. By integrating the Model Context Protocol (MCP) as a standardized tool interface, diverse agent architectures can be connected to and evaluated against the benchmark without implementation-specific modifications. A single-agent implementation demonstrates the benchmark's applicability, establishing quantitative metrics for comparison of LLM-based planning and execution approaches.
\end{abstract}


\end{frontmatter}
\section{Introduction}
Modern industrial automation systems increasingly operate in dynamic and complex environments, where changing tasks, system configurations, and heterogeneous resources require flexible and adaptive control strategies \citep{Koren.2018}. Whether in manufacturing, logistics, or multi-robot systems, such environments demand solutions that can handle variability and enable goal-directed process execution under changing conditions. 

A well-established approach for handling such complexity is \emph{AI Planning}, where planning systems automatically generate action sequences to achieve a desired goal from an initial state based on a formal environment model. Classical symbolic planning methods provide robust mechanisms for reasoning in well-structured, deterministic domains \citep{Ghallab.2004}. However, they typically require extensive manual modeling, are sensitive to incomplete or ambiguous input, and often lack flexibility to adapt to unforeseen scenarios \citep{Rogalla.2017}.

Recent advances in research on \glspl{llm} have shown strong potential for high-level reasoning, instruction following, and the generation of both structured outputs -- such as action sequences or code -- and natural-language explanations, based on unstructured input \citep[e.g.][]{ShunyuYao.2023}. 
These abilities make \glspl{llm} a promising foundation for \emph{\gls{llm} agents}, which combine the \gls{llm} with tool use and memory to enable planning and interaction with external environments \citep{Wang.2024}. Such agents can plan and execute tasks in industrial automation contexts, especially when dealing with loosely specified goals or environments that are only partially known at runtime or subject to frequent change. As a result, diverse \gls{llm} agent architectures \citep[e.g.][]{Xia.2023} have emerged. However, these approaches remain difficult to compare due to the lack of standardized benchmarks for evaluating agents on both planning and execution.

To address this gap, we introduce a benchmark based on the classic Blocksworld domain \citep{Slaney.2001} for evaluating \gls{llm}-based agents. Our benchmark provides an executable simulation environment with scenarios of varying difficulty. This well-established symbolic domain enables comparison between different \gls{llm}-based architectures as well as with classical symbolic planning methods.
To enable environment interaction, we integrate the \gls{mcp}\footnote{\url{https://modelcontextprotocol.io/specification/2025-06-18}}, which has shown promise for industrial automation \citep{VieiradaSilva.2025}, as a standardized interface enabling \glspl{llm} to discover and invoke external tools. This allows diverse agent implementations to interact with the benchmark without implementation-specific modifications, enabling systematic evaluation under consistent conditions.

Our contribution is twofold: (1) we present a structured benchmark for evaluating \gls{llm}-based planning and execution in Blocksworld; and (2) we demonstrate its applicability through the evaluation of a single-agent \gls{llm} implementation interacting with the environment via \gls{mcp}.
\section{Related Work}
\label{sec:sota}

The evaluation of agent-based systems has attracted significant attention in recent years, resulting in a wide array of benchmarks targeting various dimensions of \gls{llm}-based agentic performance.
\citeauthor{Li.21.04.2025} provide an overview of such activities and categorize benchmarks for the planning capabilities of \glspl{llm} into embodied environments, web navigation, scheduling, games and puzzles, and the automation of everyday tasks \citep{Li.21.04.2025}.
Blocksworld, as an established planning domain, belongs to the embodied environments category and has long served as a canonical testbed for symbolic reasoning~\citep{Slaney.2001}.
It forms the foundation for many benchmarks aimed at evaluating the ability of agents -- classical or \gls{llm}-based -- to generate valid plans in structured settings~\citep[e.g.][]{Valmeekam.2023, Parashar.18.02.2025}.

Blocksworld-based benchmarks -- such as PlanBench \citep{Valmeekam.2023} and Sys2Bench~\citep{Parashar.18.02.2025} -- concentrate on planning ability but use static, predefined datasets and do not integrate modular tool interfaces that would support real-time execution and adaptability.
Hence, these setups are limited in variability and lack robust provisions for evaluating plan validity, execution reliability, and the agents' capability of adaption to environmental changes.

To address the need for dynamic interaction and execution capabilities, recent benchmarks have started to incorporate tool-use capabilities via \gls{mcp}, allowing agents to interact with external APIs and services.
For instance, MCP-RADAR~\citep{Gao.22.05.2025} and MCP-Universe~\citep{Luo.20.08.2025} require \gls{llm} agents to interact with \gls{mcp} servers but focus on tool execution rather than explicit planning, limiting analysis of reasoning about dynamics and action consequences.
MCP-Bench~\citep{Wang.2025} extends this concept, focusing on multi-step tool-using tasks across diverse areas such as finance, scientific computing, or traveling.
AgentBench~\citep{Liu.2025} evaluates \gls{llm} agents across eight different environments without relying on \gls{mcp}, emphasizing multi-round interaction and long-term reasoning in scenarios ranging from operating system commands to household tasks.
While these benchmarks successfully evaluate tool-use and cross-domain workflow orchestration, they lack controlled symbolic environments where planning logic and agent reasoning remain systematically analyzable and interpretable.
Consequently, evaluating both core planning capabilities and subsequent tool-based execution of generated plans remains limited.

Despite these specific limitations, \citeauthor{Li.21.04.2025} identify broader challenges in current benchmarks that restrict their applicability to dynamic and realistic industrial automation scenarios:
They stress that static benchmarks enable \gls{llm} agents to excel by pattern-matching against familiar outputs, rather than engaging in model-based reasoning.
They highlight the absence of dynamic environments that require agents to infer and maintain internal world models, which is essential for practical deployment.
Critical gaps are especially apparent in the lack of long-horizon evaluation, insufficient mechanisms for state tracking and error correction, over-reliance on fully observable and deterministic scenarios, and the largely unimodal (text-only) nature of existing tasks.
Agents are rarely assessed on their ability to adapt plans when early errors propagate, recover from mistakes, or reason under uncertainty and incomplete information.~\citep{Li.21.04.2025} \\
As a result, the emphasis on static, text-based planning tasks that neglect execution and validation decouples agent reasoning from direct environmental interaction, reducing evaluation realism and significance.

In response to these identified gaps, our work provides several distinct advantages:
By extending Blocksworld with novel complexity dimensions -- including limited space on the table, incomplete knowledge, and constraints that limit permitted actions -- we offer a level of realism and difficulty absent from prior symbolic benchmarks.
Integration with \gls{mcp} supports dynamic tool discovery and invocation, ensuring agents interact with the environment in real-time, not only for planning, but for execution and replanning in case of failure.
In summary, our benchmark bridges the gap between static symbolic benchmarks and dynamic agent evaluation frameworks, introducing a robust \gls{mcp}-integrated environment in which both planning and execution can be rigorously tested, including conditions like incomplete information. 
\section{Blocksworld Benchmark Architecture}
\label{sec:benchmark}

\begin{figure*}[h]
\begin{center}
\includegraphics[width=\linewidth]{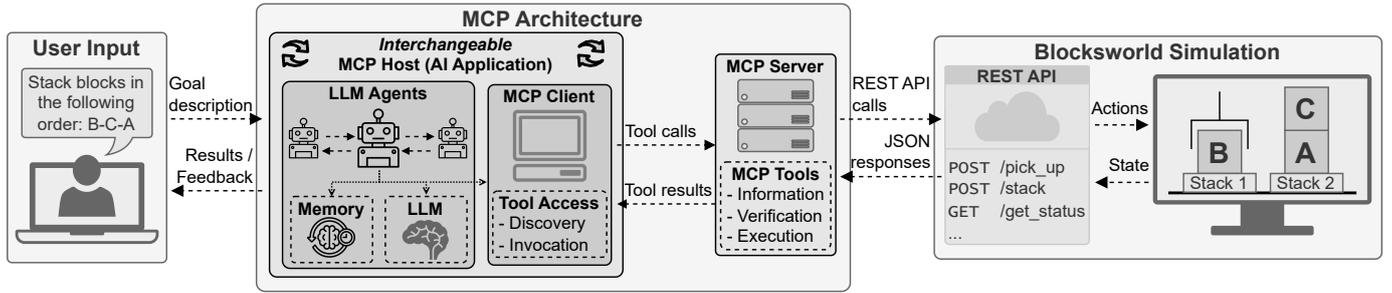}
\caption{Architecture of the Blocksworld benchmark with interaction through \gls{mcp} with interchangeable \gls{llm} agents}
\label{fig:benchmark-architecture}
\end{center}
\end{figure*}

To systematically evaluate and compare different \gls{llm} agents for planning and execution, we propose a benchmark built around the Blocksworld domain. The benchmark provides a controlled symbolic environment with well-defined constraints and adjustable complexity categories. A key design principle is modularity: the architecture enables seamless integration and interchangeability of different \gls{llm} agent implementations through the standardized \gls{mcp} interface, facilitating fair comparisons and reproducible experimentation.

The overall system architecture is illustrated in Figure~\ref{fig:benchmark-architecture}. Users provide goal descriptions in natural language, which are processed by the \gls{llm} agents within the \gls{mcp} host. The Blocksworld simulation\footnote{\url{https://github.com/hsu-aut/blocksworld_simulation}} forms the foundation, providing a symbolic environment that supports both planning and step-wise execution of actions with full state tracking and constraint validation. All simulation capabilities are exposed via a REST API, which serves a dual purpose: it enables integration of \gls{llm}-based agents while simultaneously allowing comparison with classical symbolic AI planning approaches. Traditional planners can generate action sequences and execute them through the same API interface, enabling direct performance comparisons under identical conditions.

For \gls{llm} agent integration, all REST API endpoints are wrapped as \gls{mcp} tools and exposed through an independent \gls{mcp} server\footnote{\url{https://github.com/hsu-aut/blocksworld_mcp-server}}. As shown in Figure~\ref{fig:benchmark-architecture}, the information flow follows a layered architecture: The \gls{mcp} server exposes tools in three categories (Information, Verification, and Execution -- detailed in Subsection~\ref{subsec:MCP-tools}) and communicates with the simulation through REST API calls, receiving JSON-formatted responses. The \gls{mcp} client provides tool discovery and invocation capabilities, exchanging structured tool calls and results with the \gls{mcp} server. The \gls{mcp} host can be any framework or application that implements the \gls{mcp} client and hosts the \gls{llm} agent architectures that perform the actual planning and execution. Various \gls{llm} agent architectures can be deployed, ranging from simple single-agent frameworks to complex multi-agent orchestration systems. The agents interact with the \gls{mcp} client to access simulation capabilities without requiring custom integration code for each agent implementation. Results and feedback are usually returned to the user, completing the interaction cycle. However, this depends on the specific agent implementation.

This modular design emphasizes interchangeability at the agent level: researchers can evaluate different \gls{llm} agent architectures by simply connecting their own \gls{mcp} host and agent implementation to the unchanged simulation and \gls{mcp} server. Both, the simulation and \gls{mcp} server, are provided as open-source components, ensuring transparency and extensibility. 

\subsection{Blocksworld Domain Definition}
The Blocksworld domain models an environment of movable blocks that can be manipulated by a robotic gripper according to well-defined rules. As illustrated in Figure~\ref{fig:blocksworld-example}, each planning problem is defined by an initial state and a goal state, both represented as specific block arrangements on a discrete surface. The robot can execute four classical primitive actions \citep{Slaney.2001}:

\begin{figure}[b]
\begin{center}
\includegraphics[width=\linewidth]{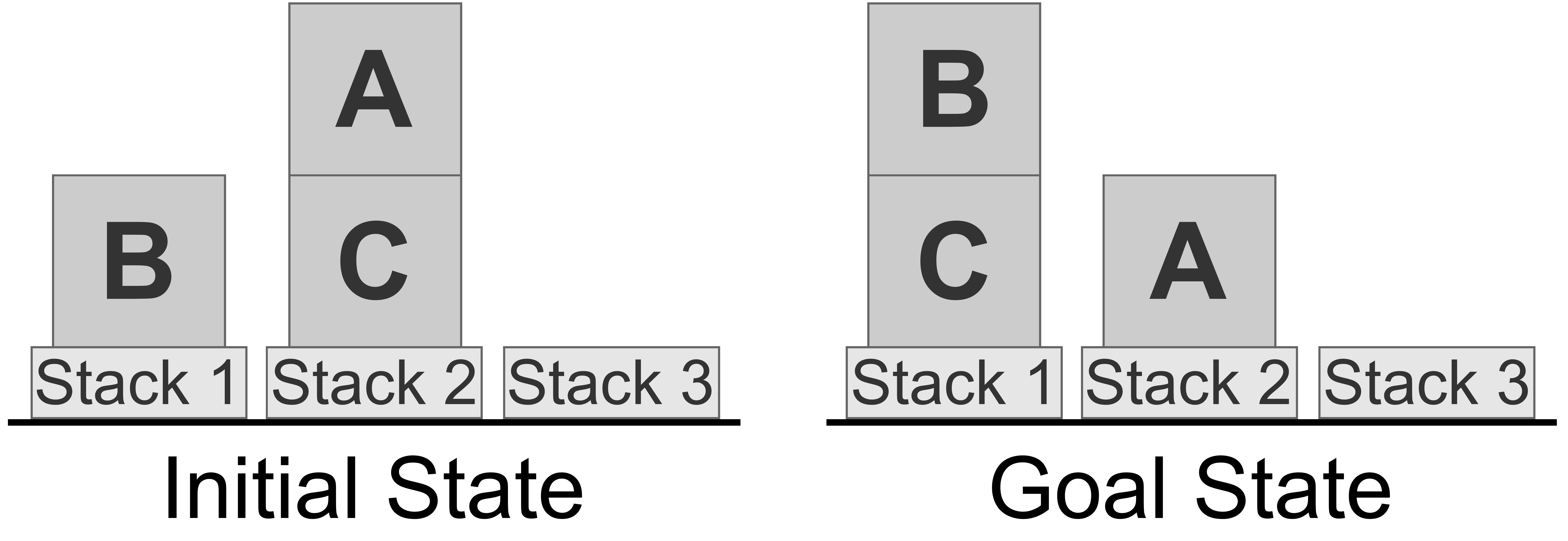} 
\caption{Blocksworld: Example of a planning problem} 
\label{fig:blocksworld-example}
\end{center}
\end{figure}

\begin{itemize}
    \item \texttt{pick\_up(block\_x)}: Grasps block x and removes it directly from the table surface.
    \item \texttt{put\_down(block\_x)}: Places the currently held block x onto an unoccupied table position.
    \item  \texttt{stack(block\_x, block\_y)}: Places the currently held block x on top of another block y.
    \item \texttt{unstack(block\_x, block\_y)}: Grasps block x and removes it from the top of another block y.
\end{itemize}

The domain is governed by the following classical constraints:

\begin{enumerate}
\item Only one block can be manipulated by the robotic gripper at a time.
\item A block can only be grasped if no other block rests on top of it.
\item Each block can have at most one block directly on top of it. 
\end{enumerate}

In our benchmark implementation, we add an additional spatial constraint that can increase the complexity of planning by limiting the available workspace: 

\begin{enumerate}
\setcounter{enumi}{3}
\item The table surface is restricted to a finite number of discrete positions, where only one stack of blocks can be placed.
\end{enumerate}

\subsection{Complexity Dimensions and Scenario Categories}
Within the benchmark, task complexity can be varied along four independent dimensions to evaluate different aspects of \gls{llm} agent capabilities. First, the \textbf{number of blocks} directly determines the state space size and the branching factor of possible action sequences.  
As in \cite{Slaney.2001}, the complexity of a scenario can also be measured by the number of \textbf{misplaced blocks} -- i.e., blocks that are not yet positioned at their destination and therefore need to be moved to achieve the goal. The greater the number of misplaced blocks, the more movements are required to achieve the desired configuration. 
Second, many goal configurations cannot be achieved through direct constructive actions alone, which introduces the dimension of \textbf{non-constructive actions}. \citeauthor{Slaney.2001} define an action as constructive if, after executing, the moved block is at its goal position and therefore does not need to be moved again to reach the goal state. Instead, agents must perform intermediate steps that temporarily move blocks away from their goal positions to create access to obstructed blocks. The ratio of such non-constructive actions to constructive actions serves as a key indicator of planning complexity. 
Third, \textbf{domain constraints} can be imposed to restrict valid actions. In particular, the \emph{block\_size} constraint introduces block sizes and permits stacking only when a smaller block is placed on top of an equal or larger one, thereby significantly increasing planning difficulty by reducing the set of valid action sequences. 
Fourth, while the classical Blocksworld assumes full state knowledge, the dimension of \textbf{incomplete information} includes scenarios where only the top two blocks of each stack are visible. This requires agents to actively query the environment or reason about hidden states, simulating real-world sensing limitations.

Based on these dimensions, benchmark scenarios are organized into five categories. Each category addresses a specific type of complexity, with varying difficulty within each category:
\begin{itemize}
    \item \textbf{Category 1 (Basic):} Basic Blocksworld rules, no non-constructive actions required, complete information, no additional constraints. 
    \item \textbf{Category 2 (With non-constructive actions):} Basic Blocksworld rules, non-constructive actions are required to solve the problem, complete information, no additional constraints. 
    \item \textbf{Category 3 (Impossible):} Unsolvable with consideration of the given constraints,  contains scenarios with different constraint sets.  
    \item \textbf{Category 4 (Additional Constraints):} The basic Blocksworld rules are extended by additional constraints regarding the allowed stacking order due to different block sizes.
    \item \textbf{Category 5 (Partial Observability):} Basic Blocksworld rules, only partial observability of the current system status.
\end{itemize}

\subsection{Simulation Environment}
The simulation is implemented in Python using \texttt{pygame}, providing a 2D visual representation of the Blocksworld environment. The architecture follows a continuous loop that processes action requests, validates them through a constraint management system, updates the internal state, and renders the current scene. Visual feedback supports both human observation and debugging of agent behavior.

\textbf{Constraint System:} All actions are validated before execution through a modular constraint management system. Three constraint sets are currently implemented: The \texttt{base} constraint set enforces classical Blocksworld rules, validating conditions such as block accessibility (top of stack), gripper state (empty or holding), and spatial availability (free table positions). The \texttt{block\_size} constraint set extends these rules with size-based stacking restrictions, ensuring that only smaller or equally sized blocks can be placed on larger ones. The well-known \emph{Towers of Hanoi} problem can also be modeled with this constraint set and is included in the provided scenarios. The \texttt{partial\_observability} constraint set ensures that only the names of the top two blocks in a stack are visible, while all other blocks are marked as unknown.
Constraint violations result in action rejection with detailed error messages in natural language, enabling agents to understand and adapt to domain restrictions.

\textbf{Robot Model:} The robotic gripper is modeled as a finite state machine that transitions through sequential states during action execution (e.g., idle, picking, holding, releasing). This state-based approach ensures realistic action execution sequences and prevents invalid state transitions. 

\textbf{Scenario Management:} The benchmark currently includes 50 predefined scenarios stored as JSON files. Each scenario specifies: (1) an initial block configuration and robot state, (2) a goal configuration with additional natural language description as possible user input, (3) the applicable constraint set, and (4) metadata including category (1--5), minimum known solution length, and resource counts (blocks, stack positions, misplaced blocks). 
The modular architecture allows researchers to easily extend the benchmark by adding custom scenarios, implementing additional constraint sets, or modifying existing rules through simple JSON configuration files, without requiring modifications to the simulation core.

\textbf{REST API:} The simulation exposes all functionality through a \texttt{flask}-based REST API with structured request validation and unified response formats. \emph{Simulation control} provides start/stop functionality with scenario selection or custom configurations. \emph{Action execution} endpoints implement the four primitive block manipulation actions with appropriate parameters. \emph{State queries} return system status, active constraint rules and available scenarios. \emph{Plan verification} supports verifying action sequences without modifying the simulation state.

\subsection{MCP Tools for LLM Agent Interaction}
\label{subsec:MCP-tools}
All simulation capabilities are exposed through an \gls{mcp} server that wraps the REST API endpoints as standardized tools. Each tool includes a natural language description of its functionality, required arguments, and expected behavior, enabling \gls{llm} agents to autonomously determine when and how to use each tool. The tools are organized into three functional categories:

\textbf{Information Tools:} The \texttt{get\_rules} tool provides a natural language description of the current problem instance, including active constraints and available actions. Rule descriptions are dynamically retrieved from the simulation, allowing a single \gls{mcp} server to support multiple scenarios with different constraint sets. This design enables generic \gls{llm} agent implementations that are not specifically tailored to Blocksworld -- agents can query the rules at runtime and adapt their behavior accordingly, rather than requiring domain-specific hardcoded knowledge. The \texttt{get\_status} tool returns the current system state as a structured JSON object, including stack configurations, block positions and properties (name, dimensions), and robot state (idle/holding, current block if any).

\textbf{Verification Tool:} The \texttt{verify\_plan} tool accepts a complete action sequence in JSON format and verifies its executability without modifying the simulation state. Each action in the sequence is checked against the current constraints and state. If the plan is verified, the tool confirms executability. If any action violates constraints, the tool identifies the first incorrect action and provides a detailed explanation of the constraint violation in natural language, enabling agents to understand and perform replanning.

\textbf{Execution Tools:} The four primitive actions (\texttt{pick\_up}, \texttt{put\_down}, \texttt{stack}, \texttt{unstack}) are exposed as individual tools, each following a uniform structure that specifies: (1) general functionality, (2) preconditions that must be satisfied, (3) effects on the environment state, (4) required input arguments, and (5) response format. For example, \texttt{stack} requires that the robot holds the block and the target block is located on top of a stack; its effect is that the held block is placed on the target and the gripper becomes free. This structured representation supports both, agent reasoning about action applicability and execution monitoring.
\section{Experimental Implementation}

To demonstrate the applicability of the benchmark and establish baseline performance metrics, we implemented a single-agent system that autonomously performs planning, verification, and execution within the Blocksworld environment, demonstrating how the benchmark enables systematic evaluation of \gls{llm}-based agents.

\subsection{Single-Agent Architecture and Configuration}

The agent is implemented as a ReAct agent in \texttt{LangGraph}, combining iterative reasoning with tool-based action execution\footnote{The complete implementation and evaluation results are available at \url{https://github.com/hsu-aut/blocksworld_single-agent}}. This architecture enables the agent to autonomously decide which tools to invoke and in what sequence, solely based on prompt-based instructions rather than a hardcoded control flow. The agent maintains unrestricted access to the complete tool suite described in Section~\ref{subsec:MCP-tools}: informational tools (\texttt{get\_rules}, \texttt{get\_status}), verification (\texttt{verify\_plan}), and execution tools (\texttt{pick\_up}, \texttt{put\_down}, \texttt{stack}, \texttt{unstack}).

The implementation is model-agnostic, supporting arbitrary \glspl{llm} through configurable API integration. For the evaluation reported here, we employed OpenAI's \textit{o3} model (snapshot \textit{o3-2025-04-16}), a reasoning-capable model that performs internal chain-of-thought processing before generating responses. This model provides a context window of 200,000 tokens and a maximum output length of 100,000 tokens, with pricing of \$2 per million input tokens and \$8 per million output tokens.

The agent's behavior is governed by a system prompt that defines a structured seven-step workflow: (1) retrieve environment rules via \texttt{get\_rules}, (2) query current state using \texttt{get\_status}, (3) generate a plan to solve the planning task, (4) verify the plan through \texttt{verify\_plan}, (5) correct incorrect steps based on verification feedback and re-verify if necessary, (6) execute the verified plan using primitive action tools, and (7) confirm goal achievement. This prompt-based control structure aims to compensate for the absence of architectural flow control present in multi-agent approaches.

The agent maintains a short-term memory that tracks the current interaction context, intermediate results, and conversation state throughout multi-step planning sessions. This memory is reset at the beginning of each new task to ensure reproducibility and unbiased problem-solving. All token usage is tracked for transparency and cost analysis, enabling measurement of both input and output token consumption per request.

\subsection{Evaluation and Results}

The single-agent system was systematically evaluated across 50 predefined benchmark scenarios spanning the five categories presented, with 10 scenarios per category. Scenarios range from 3 to 20 blocks with solution lengths of approximately 4 to 80 steps, featuring 0 to 60 non-constructive actions, up to 10 misplaced blocks, and utilizing 3 to 6 stack positions. Each scenario was executed once. Evaluation metrics included success rate (percentage of scenarios solved correctly), execution time (total time from goal specification to completion), number of planning attempts (iterations required before successful verification), token consumption (total input and output tokens), and cost (USD per scenario based on token pricing).

\begin{table}[h]
\begin{center}
\caption{Average single-agent evaluation metrics with model \textit{o3} by category}
\label{tab:single-agent-results}
\begin{tabular}{lccccc}
\hline
\textbf{Metric} & \textbf{C1} & \textbf{C2} & \textbf{C3} & \textbf{C4} & \textbf{C5} \\ 
\hline
Success Rate & 80~\% & 70~\% & 100~\% & 70~\% & 60~\%\\
Time & 76 s & 290 s & 125 s & 732 s & 676 s\\
Attempts & 1.1 & 1.7 & 1.8 & 2.2 & 3.1\\
In Tokens & 33,454 & 99,444 & 10,300 & 108,857 & 151,429\\
Out Tokens & 1,672 & 12,277 & 7,895 & 34,858 & 40,816\\
Cost &\$ 0.08 &\$ 0.30 &\$ 0.08 &\$ 0.50 &\$ 0.63\\
\hline
\end{tabular}
\end{center}
\end{table}

Table~\ref{tab:single-agent-results} summarizes the average measured values for all evaluation metrics across the five categories. The agent's results varied considerably, with success rates declining from 80~\% in category 1 to 60~\% in category 5. At category 1 (basic), tasks were completed in 76 seconds with 1.1 planning iterations on average, consuming approximately 35,100 tokens combined for input and output. Category 2 (with non-constructive actions) showed measurable degradation with a 70~\% success rate, 290 seconds execution time, 111,700 tokens consumed, and 1.7 planning attempts, indicating that initial plans frequently required correction. At category 3 (impossible), the agent correctly identified all unsolvable scenarios in on average 125 seconds with 1.8 attempts and 18,200 tokens.
Categories 4 and 5 presented the greatest challenges. Despite maintaining a 70~\% success rate, category 4 (additional constraints) required significantly longer execution times (732 seconds on average), higher token consumption (143,700), and more solution attempts (2.2). The agent showed systematic difficulties with scenarios requiring extensive non-constructive actions and occasionally misclassified solvable problems as unsolvable. Category 5 (partial observability) exhibited the poorest results across the majority of metrics: 60~\% success rate, average execution time of 676 seconds, 3.1 attempts per scenario, and approximately 192,000 tokens consumed per scenario. In category 5 scenarios, faulty tool executions occurred for the first time, in which the agent called tools with incorrect block names as argument.

These results demonstrate that while current reasoning-capable \glspl{llm} can effectively handle basic symbolic planning tasks, performance declines as the complexity of the scenario increases, especially with additional constraints such as \texttt{block\_size} or \texttt{partial\_observability}. The benchmark successfully distinguishes between complexity levels by introducing five different categories and provides quantitative metrics for systematic comparison. Additional metrics for comprehensive evaluation are available in the benchmark repository. The observed failure modes -- invalid intermediate step generation, constraint violation during execution, and premature termination -- highlight specific areas for architectural improvement in future \gls{llm} agent designs.
\section{Conclusion}

This paper introduced a systematic benchmark for evaluating \gls{llm}-based agents on planning and execution tasks using the Blocksworld domain. The benchmark provides an executable simulation environment with scenarios organized into five complexity categories and integrates \gls{mcp} as a standardized interface, enabling seamless integration and evaluation of diverse agent architectures under consistent conditions without implementation-specific modifications. This modular design supports comparison not only between different \gls{llm}-based approaches but also with classical symbolic planning methods. 
A single-agent implementation using OpenAI's \textit{o3} model demonstrated the benchmark's applicability, establishing quantitative metrics including success rate, execution time, and cost. The results showed success rates ranging from 60\% to 80\% across solvable scenarios, while achieving 100\% accuracy in identifying impossible scenarios.

The presented benchmark forms the foundation for several directions of future work. First, a systematic comparison of different \gls{llm} agent architectures for planning and execution -- including single-agent, hierarchical multi-agent, and, importantly, hybrid approaches that combine \gls{llm}-based reasoning with formal planning methods -- should be conducted to identify which architectural patterns best address the observed challenges of reliability, multi-step reasoning, and error recovery. 

Second, the benchmark scenarios should be extended to increase realism and difficulty: (1) Introduction of dynamic events such as runtime errors, changing goal configurations, or unexpected physical disturbances (e.g., collapsing stacks). (2) Additional constraints modeling real-world limitations such as block weight, material properties, or limited stacking heights. (3) Multi-robot scenarios where multiple robotic grippers coordinate their actions to achieve shared goals would introduce challenges of resource allocation. (4) More scenarios with ambiguous or incomplete initial specifications in combination with complex constraints to test agent robustness when dealing with uncertain information.

\begin{ack}
This work was supported by Siemens AG. 
\end{ack}

\section*{DECLARATION OF GENERATIVE AI AND AI-ASSISTED TECHNOLOGIES IN THE WRITING PROCESS}
During the preparation of this work the author(s) used Claude Sonnet 4.5 and GPT-4.1 in order to improve language and readability. After using this tool/service, the author(s) reviewed and edited the content as needed and take(s) full responsibility for the content of the publication.

\bibliography{ifacconf}             

\begin{thebibliography}{15}
\providecommand{\natexlab}[1]{#1}
\providecommand{\url}[1]{\texttt{#1}}
\providecommand{\urlprefix}{URL }
\expandafter\ifx\csname urlstyle\endcsname\relax
  \providecommand{\doi}[1]{doi:\discretionary{}{}{}#1}\else
  \providecommand{\doi}{doi:\discretionary{}{}{}\begingroup \urlstyle{rm}\Url}\fi

\bibitem[{Gao et~al.(2025)Gao, Xie, Zhai, Ma, and Shen}]{Gao.22.05.2025}
Gao, X., Xie, S., Zhai, J., Ma, S., and Shen, C. (2025).
\newblock {MCP-RADAR: A Multi-Dimensional Benchmark for Evaluating Tool Use Capabilities in Large Language Models}.

\bibitem[{Ghallab et~al.(2004)Ghallab, Nau, and Traverso}]{Ghallab.2004}
Ghallab, M., Nau, D.S., and Traverso, P. (2004).
\newblock \emph{{Automated planning: Theory and Practice}}.
\newblock {The Morgan Kaufmann Series in Artificial Intelligence}. {Elsevier/Morgan Kaufmann}, Amsterdam.

\bibitem[{Koren et~al.(2018)Koren, Gu, and Guo}]{Koren.2018}
Koren, Y., Gu, X., and Guo, W. (2018).
\newblock {Reconfigurable manufacturing systems: Principles, design, and future trends}.
\newblock \emph{{Frontiers of Mechanical Engineering}}, 13(2), 121--136.

\bibitem[{Li et~al.(2025)Li, Chen, Zhang, and Liu}]{Li.21.04.2025}
Li, H., Chen, Z., Zhang, J., and Liu, F. (2025).
\newblock {PLANET: A Collection of Benchmarks for Evaluating LLMs' Planning Capabilities}.

\bibitem[{Liu et~al.(2023)Liu, Yu, Zhang, Xu, Lei, Lai, Gu, Ding, Men, Yang, Zhang, Deng, Zeng, Du, Zhang, Shen, Zhang, Su, Sun, Huang, Dong, and Tang}]{Liu.2025}
Liu, X., Yu, H., Zhang, H., Xu, Y., Lei, X., Lai, H., Gu, Y., Ding, H., Men, K., Yang, K., Zhang, S., Deng, X., Zeng, A., Du, Z., Zhang, C., Shen, S., Zhang, T., Su, Y., Sun, H., Huang, M., Dong, Y., and Tang, J. (2023).
\newblock {{AgentBench}}: {{Evaluating LLMs}} as {{Agents}}.

\bibitem[{Luo et~al.(2025)Luo, Shen, Yang, Zhao, Jwalapuram, Saha, Sahoo, Savarese, Xiong, and Li}]{Luo.20.08.2025}
Luo, Z., Shen, Z., Yang, W., Zhao, Z., Jwalapuram, P., Saha, A., Sahoo, D., Savarese, S., Xiong, C., and Li, J. (2025).
\newblock {MCP-Universe: Benchmarking Large Language Models with Real-World Model Context Protocol Servers}.

\bibitem[{Parashar et~al.(2025)Parashar, Olson, Khurana, Li, Ling, Caverlee, and Ji}]{Parashar.18.02.2025}
Parashar, S., Olson, B., Khurana, S., Li, E., Ling, H., Caverlee, J., and Ji, S. (2025).
\newblock {Inference-Time Computations for LLM Reasoning and Planning: A Benchmark and Insights}.

\bibitem[{Rogalla and Niggemann(2017)}]{Rogalla.2017}
Rogalla, A. and Niggemann, O. (2017).
\newblock {Automated process planning for cyber-physical production systems}.
\newblock In \emph{{2017 22nd IEEE International Conference on Emerging Technologies and Factory Automation}}, 1--8. IEEE.

\bibitem[{Slaney and Thi{\'e}baux(2001)}]{Slaney.2001}
Slaney, J. and Thi{\'e}baux, S. (2001).
\newblock {Blocks World revisited}.
\newblock \emph{{Artificial Intelligence}}, 125(1-2), 119--153.

\bibitem[{Valmeekam et~al.(2023)Valmeekam, Maquez, Olmo, Sreedharan, and Kambhampati}]{Valmeekam.2023}
Valmeekam, K., Maquez, M., Olmo, A., Sreedharan, S., and Kambhampati, S. (2023).
\newblock {PlanBench: An Extensible Benchmark for Evaluating Large Language Models on Planning and Reasoning about Change}.
\newblock In \emph{37th Conference on Neural Information Processing Systems (NeurIPS 2023)}, Advances in neural information processing systems. {Curan Associates Inc}, Red Hook, NY.

\bibitem[{Vieira~da Silva et~al.(2025)Vieira~da Silva, K{\"o}cher, and Gehlhoff}]{VieiradaSilva.2025}
Vieira~da Silva, L.M., K{\"o}cher, A., and Gehlhoff, F. (2025).
\newblock {Beyond Formal Semantics for Capabilities and Skills: Model Context Protocol in Manufacturing}.
\newblock In \emph{{2025 IEEE 30th International Conference on Emerging Technologies and Factory Automation}}, 1--4. IEEE.

\bibitem[{Wang et~al.(2024)Wang, Ma, Feng, Zhang, Yang, Zhang, Chen, Tang, Chen, Lin, Zhao, Wei, and Wen}]{Wang.2024}
Wang, L., Ma, C., Feng, X., Zhang, Z., Yang, H., Zhang, J., Chen, Z., Tang, J., Chen, X., Lin, Y., Zhao, W.X., Wei, Z., and Wen, J. (2024).
\newblock {A survey on large language model based autonomous agents}.
\newblock \emph{{Frontiers of Computer Science}}, 18(6).

\bibitem[{Wang et~al.(2025)Wang, Chang, Patel, Biju, Wu, Liu, Ding, Rezazadeh, Shah, Bao, and Siow}]{Wang.2025}
Wang, Z., Chang, Q., Patel, H., Biju, S., Wu, C.E., Liu, Q., Ding, A., Rezazadeh, A., Shah, A., Bao, Y., and Siow, E. (2025).
\newblock {{MCP-Bench}}: {{Benchmarking Tool-Using LLM Agents}} with {{Complex Real-World Tasks}} via {{MCP Servers}}.

\bibitem[{Xia et~al.(2023)Xia, Shenoy, Jazdi, and Weyrich}]{Xia.2023}
Xia, Y., Shenoy, M., Jazdi, N., and Weyrich, M. (2023).
\newblock {Towards autonomous system: flexible modular production system enhanced with large language model agents}.
\newblock In \emph{{2023 IEEE 28th International Conference on Emerging Technologies and Factory Automation}}, 1--8. IEEE.

\bibitem[{Yao et~al.(2023)Yao, Zhao, Yu, Du, Shafran, Narasimhan, and Cao}]{ShunyuYao.2023}
Yao, S., Zhao, J., Yu, D., Du, N., Shafran, I., Narasimhan, K.R., and Cao, Y. (2023).
\newblock {ReAct: Synergizing Reasoning and Acting in Language Models}.
\newblock In \emph{{The Eleventh International Conference on Learning Representations}}.

\end{thebibliography}
\end{document}